\documentclass[10pt,twocolumn,letterpaper]{article}

\usepackage{iccv}
\usepackage{times}
\usepackage{epsfig}
\usepackage{graphicx}
\usepackage{amsmath}
\usepackage{amssymb}
\usepackage{enumitem}
\usepackage{multirow}
\usepackage{boldline}
\usepackage{bm}
\usepackage{mathtools, nccmath}
\usepackage{algorithm,algpseudocode}
\usepackage{algpseudocode}
\usepackage{bbm}

\usepackage[pagebackref=true,breaklinks=true,letterpaper=true,colorlinks,bookmarks=false]{hyperref}

\newcommand{\myparagraph}[1]{\vspace{2pt}\noindent{\bf{#1}}}
\newcommand{\remark}[1]{\vspace{2pt}\noindent{\em{#1}}}
\iccvfinalcopy 

\def\iccvPaperID{1856} 

\ificcvfinal\fi

\begin{document}

\title{Where and When: Space-Time Attention for Audio-Visual Explanations}

\author{Yanbei Chen\textsuperscript{1}, \hspace{3pt} Thomas Hummel\textsuperscript{1}, \hspace{3pt} A. Sophia Koepke\textsuperscript{1}, \hspace{3pt} Zeynep Akata\textsuperscript{1,2,3} \\
{\small
\textsuperscript{1}University of T{\"u}bingen \hspace{2pt}
\textsuperscript{2}MPI for Informatics \hspace{2pt} 
\textsuperscript{3}MPI for Intelligent Systems} \\
{\small \{yanbei.chen, thomas.hummel, a-sophia.koepke, zeynep.akata\}@uni-tuebingen.de}
}

\maketitle
\ificcvfinal\thispagestyle{empty}\fi

\begin{abstract}
Explaining the decision of a multi-modal decision maker requires to determine the evidence from both modalities. Recent advances in XAI provide explanations for models trained on still images. However, when it comes to modeling multiple sensory modalities in a dynamic world, it remains underexplored how to demystify the mysterious dynamics of a complex multi-modal model. In this work, we take a crucial step forward and explore learnable explanations for audio-visual  recognition. Specifically, we propose a novel space-time attention network that uncovers the synergistic dynamics of audio and visual data over both space and time. Our model is capable of predicting the audio-visual video events, while justifying its decision by localizing where the relevant visual cues appear, and when the predicted sounds occur in videos. We benchmark our model on three audio-visual video event datasets, comparing extensively to multiple recent multi-modal representation learners and intrinsic explanation models. Experimental results demonstrate the clear superior performance of our model over the existing methods on audio-visual video event recognition. Moreover, we conduct an in-depth study to analyze the explainability of our model based on robustness analysis via perturbation tests and pointing games using human annotations.

\end{abstract}

\section{Introduction}

\begin{figure}[!t]
\includegraphics[width=0.478\textwidth]{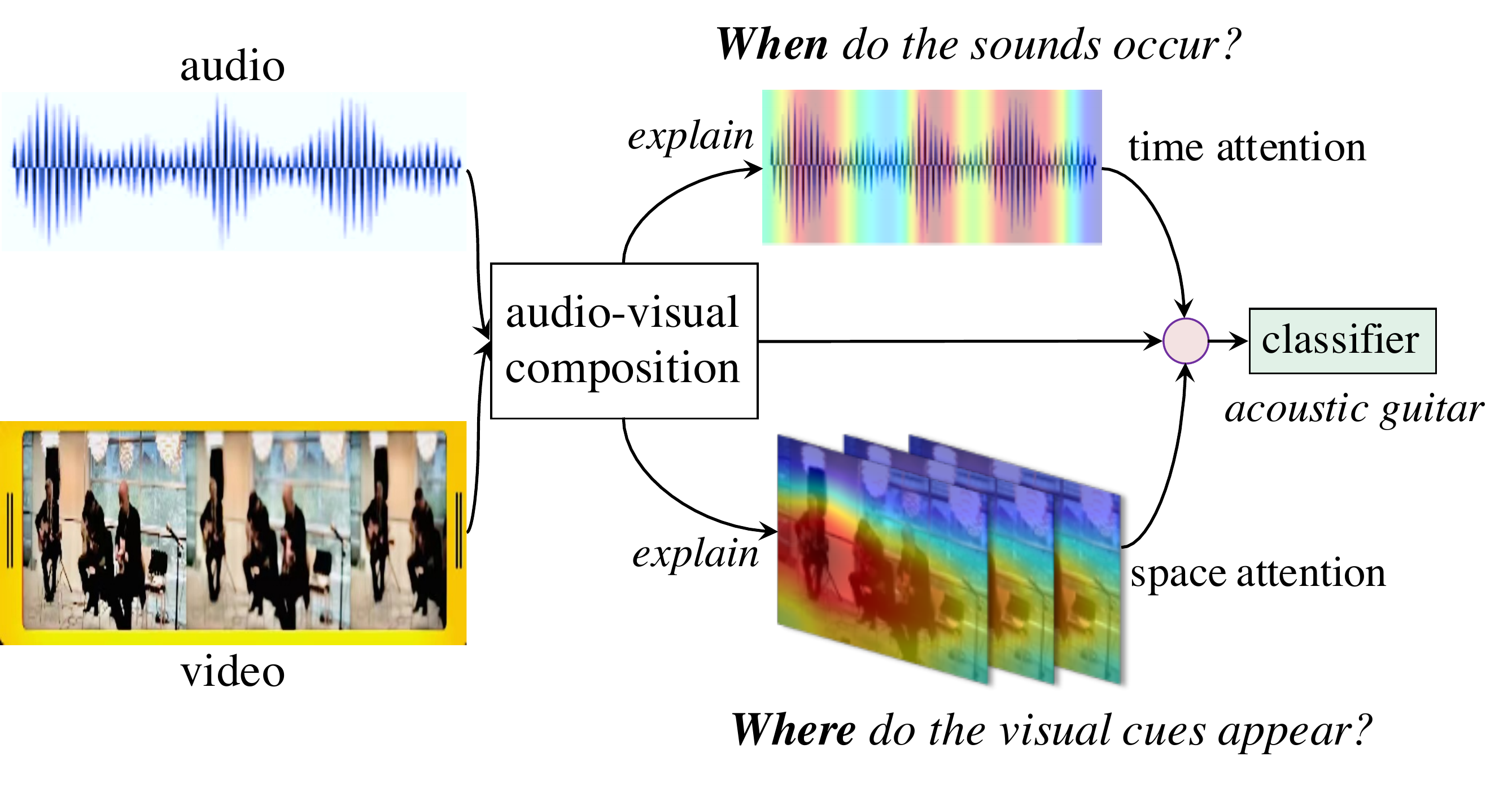}
\caption{
For a given video, our space-time attention network (STAN) predicts the audio-visual event and provides audio-visual explanations, which pinpoint {\em where} the discriminative visual cues appear in the video and {\em when} the predicted sounds occur. 
}
\label{fig:teaser}
\end{figure}

A real-world event is often perceived and interpreted by processing information from various sensory modalities, such as audio and vision \cite{ernst2004merging}. 
For instance, the video event of {\em playing acoustic guitar} can be recognized because of the presence of a musician playing the guitar in the scene (visual cues) and the sound of the guitar (audio cues).
In fact, perceiving the audio and visual cues simultaneously makes it easier to distinguish acoustically or visually similar events such as playing acoustic guitar and playing mandolin. The synergy of audio and visual modalities has also been shown to be beneficial for learning more powerful multi-modal representations for video recognition. Recently, a line of works shows that integrating the audio and visual data by multi-modal learning can greatly boost the model performance, e.g. for recognizing human action \cite{wang2020makes}, speech \cite{afouras2018deep}, or sound events \cite{fayek2020large} in videos.  

Although existing works have discovered the strength of multi-modal networks for video recognition, 
understanding how different modalities are composed and utilized for model predictions remains an unresolved challenge. Inspired by the recent advances in explainable AI (XAI) \cite{chattopadhay2018grad,hendricks2018grounding,xu2020explainable,liu2020towards,liznerski2020explainable}, our goal in this work is to uncover the underlying rationale for the audio-visual model predictions. Instead of simply fusing the two modalities, we introduce an intrinsically explainable audio-visual model that can predict the audio-visual video events, while explaining how each data modality contributes to the model predictions. 

To achieve the aforementioned intrinsic explainability in audio-visual learning, we need to tackle several 
challenges: (i) the audio and visual information could vary drastically over space and time; (ii) obtaining desired audio-visual recognition and explanation both require learning good representations. 
We tackle these challenges via a unified audio-visual model built with an explainable space-time attention mechanism (Figure \ref{fig:teaser}). 
In contrast to most existing explainable models that operate on the static images and texts \cite{selvaraju2017grad,sundararajan2017axiomatic,park2018multimodal,zellers2019recognition}, our model learns to provide explanations for the audio and visual data over space and time. 
In other words, unlike visual recognition on still images, it is crucial to simultaneously reason on the spatial and temporal dynamics across different modalities for audio-visual video recognition. 
As a pioneer in exploring explanations for audio-visual learning, 
our model is designed to predict the audio-visual video events and offer explanations that localize {where} the related visual cues appear in space and {when} the predicted sounds occur along time.  

Our contributions are as follows. 
(1) We propose a novel explainable space-time attention network (STAN) which uncovers the underlying spatial and temporal dynamics to justify how the audio and visual modalities are utilized for recognizing an audio-visual event. 
(2) We establish a comprehensive benchmark on multiple audio-visual event datasets, comparing our model to multiple recent multi-modal models and intrinsic explanation models. Experimental results show that our model achieves superior performance on audio-visual event recognition. 
(3) We provide an insightful study to analyze the explainablity of our model.  
We show that our model serves as a good proxy for model explanations and 
offers human-interpretable explanations. 

\section{Related Work}

\myparagraph{Visual Explanations.} 
A group of existing works provide visual explanations (e.g., a saliency or attention map) that 
uncover the underlying focus of deep neural networks in a visual recognition task \cite{simonyan2013deep,zeiler2014visualizing,mahendran2015understanding,zhou2016learning,selvaraju2017grad,selvaraju2017grad,smilkov2017smoothgrad,sundararajan2017axiomatic,chattopadhay2018grad,zhang2018top,fukui2019attention,liznerski2020explainable,liu2020towards}. 
Existing visual explanation models can be grouped into three families: CNN visualization \cite{zeiler2014visualizing,mahendran2015understanding}, gradient-based post-hoc explanations \cite{selvaraju2017grad,smilkov2017smoothgrad,chattopadhay2018grad,sundararajan2017axiomatic}, 
and response-based intrinsic explanations \cite{zhou2016learning,fukui2019attention}.  
While the former two families obtain visualizations without making any architectural change, the latter one incorporates visual explanations as an intrinsic component in the model architecture. 
The gradient-based visual explanation models (such as Grad-CAM \cite{selvaraju2017grad} and Integrated Gradients \cite{sundararajan2017axiomatic}) utilize the back-propagated gradients of a pre-trained network to derive an activation or saliency map for identifying the image regions or pixels that most contribute to the model's decisions. 
For response-based visual explanation models (e.g., CAM \cite{zhou2016learning}, ABN \cite{fukui2019attention} and one-class explanation models \cite{liu2020towards,liznerski2020explainable}), an activation or attention map is designed as an intrinsic property of the model to provide transparent explanations, which often yield an interpretable heatmap on the image to highlight the decisive regions. In a similar spirit as CAM and ABN, our proposed model is also an intrinsic explanation model with a learnable class activation mechanism to localize the salient regions. However, we stretch this mechanism to model the multi-modal audio-visual data in videos, and propose a novel explainable space-time attention mechanism that jointly learns class activation maps over space and class activation values along time. 

\myparagraph{Multi-Modal Explanations.}  
Another recent line of works offers multi-modal explanations for deep neural networks via generating human-interpretable justifications \cite{hendricks2016generating,park2018multimodal,kim2018textual,hendricks2018grounding,zellers2019recognition,kanehira2019multimodal,xu2020explainable}, such as text explanations that explain a model's decisions by generating human-readable words or natural language sentences \cite{kim2018textual}, or other cues like grounding bounding boxes that point to the important visual regions \cite{kanehira2019multimodal}. 
In general, these explanation models are trained in a multi-task learning framework by jointly optimizing a primary task objective and an auxiliary explanation objective. 
Among these works, several of them explore to learn from multi-modal visual and textual data and generate text explanations to justify the answers for visual question answering \cite{park2018multimodal} or visual commonsense reasoning \cite{zellers2019recognition}. Similarly, we also explore multi-modal data, but in the task of audio-visual event recognition that especially requires the synergistic space-time understanding of two data modalities. 
Although text explanations or bounding boxes are human-friendly, they often require manual annotations to provide supervision on explanations during training. 
To avoid using expensive annotations,
we propose a space-time attention mechanism to explain the spatial and temporal dynamics for recognizing a video event, without using any additional supervision on explanations. 

\myparagraph{Audio-Visual Learning.} 
Integrating the audio and visual modalities for multi-modal representation learning can benefit a wide variety of tasks, such as 
lip reading \cite{afouras2020asr}, 
speech recognition \cite{afouras2018deep},
speech separation \cite{ephrat2018looking}, 
emotion recognition \cite{noroozi2017audio,albanie2018emotion}, 
sound source separation \cite{gao2019co,afouras2020self}, 
sound localization \cite{tian2018audio,wu2019dual,gan2019self}, 
action recognition \cite{wang2020makes,xiao2020audiovisual,chen2021distilling}, 
sound recognition \cite{fayek2020large}, 
audio-driven image synthesis \cite{wiles2018x2face,jamaludin2019you,wang2020mead}, 
and visual-driven audio synthesis \cite{zhou2019vision,gao20192}. 
A line of works has also explored various ways to localize the visual regions related to sounds
\cite{arandjelovic2018objects,zhao2018sound,zhao2019sound,afouras2020self}, or to localize the sounds \cite{tian2018audio,wu2019dual}. In this work, we consider to jointly localize the audio and visual aspects over space and time. However, rather than simply fusing the two modalities for learning an audio-visual recognition model \cite{wang2020makes,fayek2020large}, we introduce explainability into audio-visual learning. While existing works on localizing sound sources \cite{wu2019dual,gan2019self} or sound source separation \cite{gao2019co,afouras2020self} focus mainly on the audio aspect and do not consider explanability, our work considers both recognition and explanations on audio and visual modalities simultaneously. Moreover, we evaluate our model under the explanation evaluation criteria to study its underlying connections to model explanations and human explanations. 

\section{Space-Time Attention Network (STAN)}
\begin{figure*}[!t]
\includegraphics[width=0.99\textwidth]{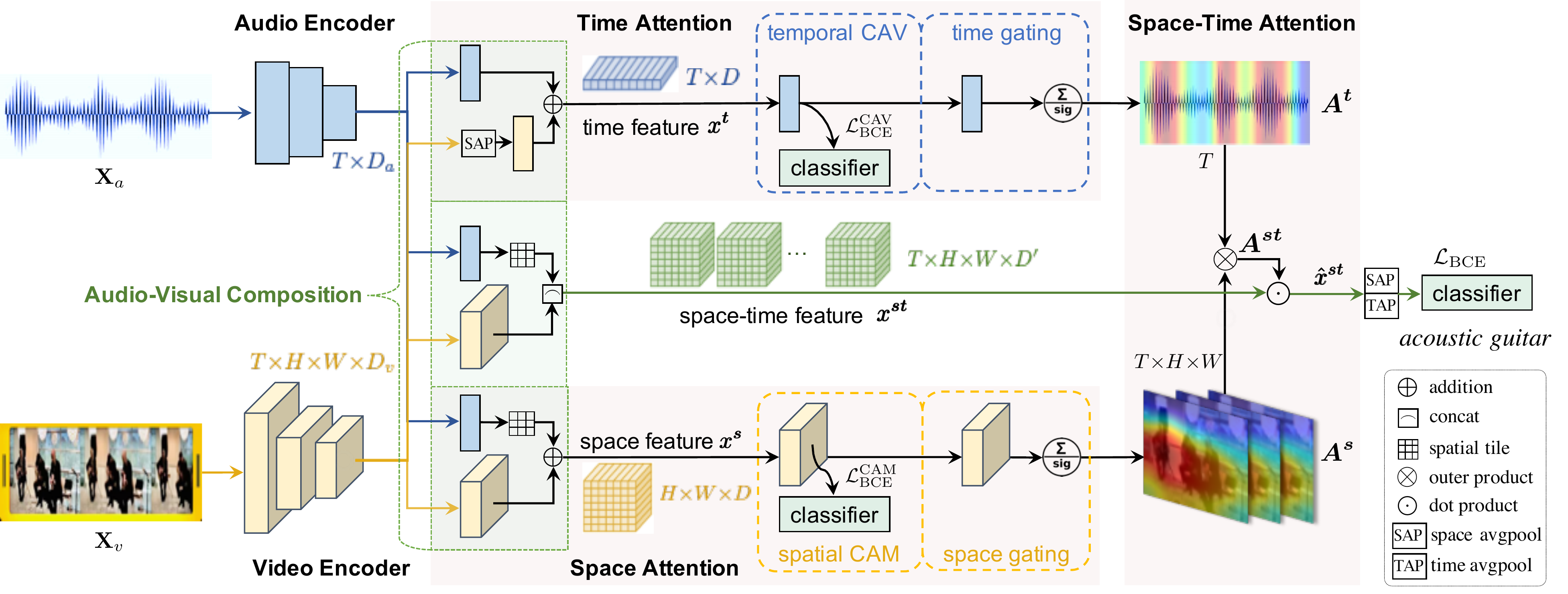}
\caption{ 
Given audio and visual data from a video, STAN first composes the audio and visual features from the audio and video encoders (Section \ref{sec:compose}), and learns space-time attention (Section \ref{sec:explain}), constrained by a learning objective for three classifiers (Section \ref{sec:optimization}). 
Once trained, STAN can recognize the audio-visual events and offer explanations on the audio-visual data. 
While the space attention tells {\em where} the salient visual cues appear in space, the time attention tells {\em when} the sounds occur along time. 
CAV/CAM: class activation values/maps. 
}
\label{fig:model}
\end{figure*}

Given a video {consisting of} 
audio and image frames, our task is to recognize the audio-visual video events, while being capable of explaining {\em where} (i.e., space) and {\em when} (i.e., time) are decisive for recognition. 
 For instance, to predict the audio-visual event of {\em playing acoustic guitar}, an audio-visual recognition model is expected to ``see'' where the musicians play the string instruments in the video, and ``hear'' when the sounds of string music occur in the audio track. 
To this end, we propose a unified audio-visual recognition model with an explainable space-time attention mechanism. 
As Figure \ref{fig:model} shows, the attention mechanism is decomposed and learned separately along the space and time dimensions, conditioning on the audio-visual features.
By design, the space-time attention can pinpoint the salient visual and audio cues, and selectively activate the space-time features to learn explainable representations for video event recognition using  heterogeneous audio and visual data. 

\subsection{Representing Audio and Visual Modalities}
\label{sec:compose}

We refer to the video dataset as $\{\textbf{X}_{a[i]}, \textbf{X}_{v[i]}, y_{[i]}\}_{i=1}^{N}$, 
where {$\textbf{X}_{a}$ denotes the audio, $\textbf{X}_{v}$ is the video}, $y \in \mathbb{R}^{{K}}$ is the corresponding video event label and $K$ is the number of classes. To learn from the fine-grained information over space and time, $\textbf{X}_{a}, \textbf{X}_{v}$ are decomposed {in}to a sequence of $T$ segments, where $T$ is the total time length. 
Each audio or video segment is of a fixed time length (e.g. 1 second). 
Next, we detail how each modality is represented and how they are composed for audio-visual representation learning. 

\myparagraph{Audio Encoder.} 
An audio track $\textbf{X}_{a}$ is represented by a sequence of audio features, i.e., $\bm{a} = \{a_{1}, \dots, a_{t}, \dots, a_{T}\}$. Each audio segment is a $D_a$-dimensional embedding $a_{t}$ extracted from the log-mel spectrogram by an audio encoder. 

\myparagraph{Video Encoder.} 
To encode the space-time visual content, the video $\textbf{X}_{v}$ is represented {by} a sequence of visual features, i.e., $\bm{v} = \{v_{1}, ..., v_{t}, ..., v_{T}\}$. For each visual segment, a ResNet \cite{he2016deep} is used to extract a $(H \times W \times D_v)$-dimensional feature map $v_t$, where $H \times W$ denotes its spatial dimensions. 
As each visual segment could contain multiple image frames, 
temporal average pooling is applied to aggregate all the frame-wise image features and derive one feature map per segment. Thus, each video {$\textbf{X}_{v}$} is encoded by a space-time visual feature tensor $\bm{v} \in \mathbb{R}^{T \times H \times W \times D_v}$.

\myparagraph{Audio-Visual Composition.} 
As audio and visual modalities encode heterogeneous and complementary information, we propose to compose {the} two modalities. 
To {achieve} this aim, an intuitive strategy {would be} to add, average or concatenate the audio and visual features, which however, may not be directly applicable due to {a} mismatch in feature dimensions across modalities. 
Hence, we apply linear transformations to project the audio and visual feature tensors to the size of $T \times D$ and $T \times H \times W \times D$, 
followed by tiling the audio features spatially to kept the audio and visual features in an identical size of $T \times H \times W \times D$. 
Formally, the linear transformations can be written as: 
\begin{equation}
\begin{aligned}
\hat{a}_{t} &= \mathcal{F}_{\text{tile}}(\mathcal{F}_{\text{MLP}}^{st}([a_t, \bar{a}]; W_a^{st})), \\
\hat{v}_{t} &= \mathcal{F}_{\text{conv}}^{st}(v_t; W_v^{st}),
\end{aligned}
\label{eq:linear}
\end{equation}
where $\mathcal{F}_{\text{MLP}}^{st}(\cdot; W_a^{st})$ is an MLP; $\mathcal{F}_{\text{tile}}(\cdot)$ is a spatial tiling operator; $\mathcal{F}_{\text{conv}}^{st}(\cdot; W_v^{st})$ is a convolutional layer. 
To take the global audio information into account, 
each audio segment is represented by concatenating its feature $a_t$ and the temporal average pooling feature of all audio segment features, i.e., $\bar{a} = \frac{1}{T}\sum_{t=1}^{T}a_{t}$. 
The features of the audio and video sequences can be referred as 
$\bm{\hat{a}}$ and $\bm{\hat{v}}$, where 
$\bm{\hat{a}}=\{\hat{a}_{1}, ... \hat{a}_{t}, ..., \hat{a}_{T}\}^{T \times H \times W \times D}$ 
and 
$\bm{\hat{v}}=\{\hat{v}_{1}, ... \hat{v}_{t}, ..., \hat{v}_{T}\}^{T \times H \times W \times D}$. 
To compose the audio and visual features $\bm{\hat{a}}$ and $\bm{\hat{v}}$, 
simple addition or concatenation can be applied to derive a compositional space-time feature tensor $\bm{x}^{\bm{st}} \in \mathbb{R}^{T \times H \times W \times D'}$ which encodes the audio-visual data. 
We concatenate $\bm{\hat{a}}$ and $\bm{\hat{v}}$ to get the space-time feature.

\subsection{Learning to Explain Where and When}
\label{sec:explain}

To demystify how the audio and visual modalities are composed for recognizing a video event, our space-time attention mechanism is learned to explain {\em where} the visual cues appear in the video and {\em when} the sounds {occur} in the audio. More precisely, the space and time attentions are learned separately and then integrated to obtain attention-weighted space-time features. Since the audio and visual modalities are both important for recognition, the space and time attentions are learned upon the space features and time features respectively, as elaborated in the following. 

\myparagraph{Space Attention.} 
To explain the {\em where} dynamics in space, we first derive the space features, followed by learning the space attention. Specifically, the space representations are obtained using Eq. \eqref{eq:linear}, which gives an audio-visual space-time feature tensor $\bm{x}^{\bm{s}} = \{x_{1}^{\bm{s}}, ..., x_{t}^{\bm{s}}, ... x_{T}^{\bm{s}} \}$. 
To learn the space attention, 
spatial class activation maps (CAM) \cite{zhou2016learning} are learned for each individual space feature tensor ${x}_{t}^{\bm{s}} \in \mathbb{R}^{H \times W \times D}$, followed by a learnable space gating function:  
\begin{equation}
\begin{aligned}
& {A}_{t}^{\bm{s}} = \mathcal{F}_{\text{space-gate}}(\mathcal{F}_{\text{CAM}}({x}_{t}^{\bm{s}}; W_{s}); W_{sa}), \\
& \text{with} \ 
\mathcal{F}_{\text{CAM}}({x}_{t}^{\bm{s}}; W_{s}) = M_t = W_{s} * {x}_{t}^{\bm{s}},
\end{aligned}
\label{eq:space}
\end{equation}
where $*$ denotes 2D convolution;  
$\mathcal{F}_{\text{CAM}}(\cdot; W_{s})$ learns the 
class activation maps for a space feature tensor $x_{t}^{\bm{s}}$ at time step $t$: $M_t \in \mathbb{R}^{H \times W \times K}$. 
$\mathcal{F}_{\text{space-gate}}(\cdot; W_{sa})$ is a space gating function with a convolutional layer and a sigmoid function, which maps the class activation maps to a space attention map ${A}_{t}^{\bm{s}} \in \{0,1\}^{H \times W}$. 
As ${A}_{t}^{\bm{s}}$ is learned upon the spatial CAM, 
it summarizes how the model arrives at its decision in space. As Figure \ref{fig:model} shows, a space attention map ${A}_{t}^{\bm{s}}$ 
essentially pinpoints the most discriminative visual regions. For a video, the overall space attention is written as $\bm{A}^{\bm{s}} = \{{A}_{1}^{\bm{s}}, ..., {A}_{t}^{\bm{s}}, ...,{A}_{T}^{\bm{s}} \}^{T \times H \times W}$, which includes $T$ space attention maps. 

\myparagraph{Time Attention.} 
Although the space attention can explain the visual dynamics, it does not tell when {the} sounds {occur}. The latter aspect is however essential to explain how the audio contributes to {the} recognition of an audio-visual event. To achieve this, we propose to first derive the time features, followed by learning the time attention. In {the} time dimension, the spatial information is no longer important; thus, we rewrite Eq. \eqref{eq:linear} to learn the time representations by linear transformations on the audio and visual features: 
\begin{equation}
\begin{aligned}
\hat{a}_{t} &= \mathcal{F}_{\text{MLP}}^{t}([a_t, \bar{a}]; W_a^{t}), \\
\hat{v}_{t} &= \mathcal{F}_{\text{MLP}}^{t}(\bar{v}_t; W_v^{t}),
\end{aligned}
\label{eq:lineartime}
\end{equation}
where $\mathcal{F}_{\text{MLP}}^{t}(\cdot;W_a^{t}), \mathcal{F}_{\text{MLP}}^{t}(\cdot;W_v^{t})$ are MLP layers. 
To discard the spatial information, $\bar{v}_t \in \mathbb{R}^{T \times D_v}$ is derived by spatial average pooling on the original visual feature map ${v}_t$. 
Given Eq. \eqref{eq:lineartime}, 
the audio and visual time features are kept in an identical size, where   
$\bm{\hat{a}}^{\bm{t}}=\{\hat{a}_{1}, ... \hat{a}_{t}, ..., \hat{a}_{T}\}^{T \times D}$ 
and 
$\bm{\hat{v}}^{\bm{t}}=\{\hat{v}_{1}, ... \hat{v}_{t}, ..., \hat{v}_{T}\}^{T \times D}$. 
Thus, $\bm{\hat{a}}^{\bm{t}}$ and $\bm{\hat{v}}^{\bm{t}}$ can be composed by addition to obtain a compositional time feature tensor: $x^{\bm{t}} = \{x_{1}^{\bm{t}}, ..., x_{t}^{\bm{t}}, ..., x_{T}^{\bm{t}}\}^{T \times D}$. 

To learn an explainable time attention in a similar {fashion} as {the} space attention (Eq. \eqref{eq:space}), we propose to first learn the temporal class activation values (CAV) per time step feature $x_{\bm{t}_t}$, followed {by} a learnable time gating function:
\begin{equation}
\begin{aligned}
& {A}_{t}^{\bm{t}} = \mathcal{F}_{\text{time-gate}}(\mathcal{F}_{\text{CAV}}(\bm{x}_{t}^{\bm{t}}; W_{t}); W_{ta}), 
\\
& \text{with} \ 
\mathcal{F}_{\text{CAV}}(\bm{x}_{t}^{\bm{t}}; W_{t}) = V_t = W_t \bm{x}_{t}^{\bm{t}},
\end{aligned}
\label{eq:time}
\end{equation}
where $\mathcal{F}_{\text{CAV}}(\cdot; W_{t})$ is an MLP that learns the class activation values for the time feature $\bm{x}_{t}^{\bm{t}}$ at time $t$; $V_t \in \mathbb{R}^{K}$ denotes the class activation values. 
$\mathcal{F}_{\text{time-gate}}(\cdot; W_{ta})$ is a time gating function with an  MLP and a sigmoid function, which maps the class activation values to an attention scalar value per time step, i.e., ${A}_{t}^{\bm{t}} {\in} \{0,1\}$. For a video, the time attention is written as 
$\bm{A}^{\bm{t}} = \{ {A}_{1}^{\bm{t}}, ..., {A}_{t}^{\bm{t}}, ..., {A}_{T}^{\bm{t}} \}^{T}$. 
$\bm{A}^{\bm{t}}$ has higher values when {relevant} sounds {occur} (see Figure \ref{fig:model}). 

\remark{Remark.} 
It is worth noting that the temporal CAV differs from the spatial CAM. While CAM learns an activation map per class per space tensor, CAV learns an activation value per class per time step. However, both CAM and CAV learn the per-class relevance scores at a specific spatial or temporal location, which are also constrained by classification losses (detailed in Section \ref{sec:optimization}). 
Thus, CAM and CAV can both offer class activation scores to learn the attention maps that indicate the feature importance over space and time. 

\myparagraph{Space-Time Attention.} 
Given the space and time attention tensors ${A}^{\bm{s}}$, ${A}^{\bm{t}}$, we 
integrate the two attention tensors by outer product, {resulting in} a space-time attention tensor:
\begin{equation}
\begin{aligned}
\bm{A}^{\bm{st}} &= \bm{A}^{\bm{s}} \otimes \bm{A}^{\bm{t}},
\end{aligned}
\label{eq:spacetime}
\end{equation}
where $\bm{A}^{\bm{st}} \in \{0,1\}^{T \times H \times W}$ is a holistic attention tensor that tells how the space-time dynamics contribute to audio-visual representation learning. For each video, ${A}^{\bm{st}}$ {operates} on its audio-visual space-time feature tensor $\bm{x}^{\bm{st}}$ (Eq. \eqref{eq:linear}) as follows. 
\begin{equation}
\begin{aligned}
\bm{\hat{x}}^{\bm{st}} &=  \bm{x}^{\bm{st}} \odot \bm{A}^{\bm{st}},
\end{aligned}
\label{eq:feature}
\end{equation}
where $\odot$ is the element-wise product of two tensors. 
The new space-time feature tensor $\bm{\hat{x}}^{\bm{st}}$ (reweighted by attention tensor ${A}^{\bm{st}}$) is further passed towards the final classifier for predicting the video events, as detailed in the next section.

\subsection{Learning Objective}
\label{sec:optimization}

Given the attention-weighted audio-visual feature (Eq. \eqref{eq:feature}), 
the STAN model is trained to predict {the} {video} event{s}. 
Each audio-visual feature tensor $\bm{\hat{x}}^{\bm{st}} \in \mathbb{R}^{T \times H \times W \times D'}$ is first passed through spatial average pooling and temporal average pooling to obtain a $D'$-dimensional feature embedding, followed by a fully-connected layer with sigmoid activation to obtain the final probabilistic predictions $p$ for the video. 
As each video may be tagged with more than one event label, we cast this recognition task as a multi-label multi-class classification problem. 
Hence, STAN can be optimized with a multi-label binary cross-entropy loss: 
\begin{equation}
\begin{aligned}
\mathcal{L}_{\text{BCE}} &= 
\sum_{j=1}^{K}y(j)\text{log} p(j) + (1{-}y(j))\text{log} (1{-}p(j)),
\end{aligned}
\label{eq:CE}
\end{equation}
where $y$ is the groundtruth audio-visual event label of the video. 
In theory, the CAM and CAV (from Eq. \eqref{eq:space} and Eq. \eqref{eq:time}) both learn the class activations. {Therefore,} we constrain the learning of CAM and CAV by mapping their average pooling features to the class predictions by adding the classifier upon CAM and CAV, 
using the same loss in Eq.~\eqref{eq:CE}. 
In particular, for CAM, we perform spatial average pooling and temporal average pooling of all the class activation maps in the full video; while for CAV, we perform temporal average pooling of all the class activation values. 
The overall learning objective for STAN can be written as: 
\begin{equation}
\begin{aligned}
\mathcal{L}_{\text{STAN}} &= 
\mathcal{L}_{\text{BCE}} + 
\mathcal{L}_{\text{BCE}}^{\text{CAM}} + \mathcal{L}_{\text{BCE}}^{\text{CAV}}.
\end{aligned}
\label{eq:all}
\end{equation}
As there are three classifiers, STAN is trained {with} $\mathcal{L}_{\text{STAN}}$ as a multi-task model. To simplify model optimization, we use pre-trained audio and video encoders and train only the space-time attention modules and classifiers, while keeping the encoders {\em frozen} during training. At test time, the final classifier (constrained by $\mathcal{L}_{\text{BCE}}$) is used for model inference.

\section{Experiments}

We first detail the experimental setup (Section \ref{sec:setup}), 
and evaluate on audio-visual recognition (Section \ref{sec:recognition}). 
Finally, we analyze the explainability of our model (Section \ref{sec:analysis}). 

\subsection{Experimental Setup}
\label{sec:setup}

\myparagraph{Datasets.} 
We evaluate on three audio-visual event datasets. 
First, {AVE} \cite{tian2018audio} is an audio-visual event dataset, including 4,143 videos of 28 video event categories (e.g., {\em motorcycle} and {\em violin}). Most videos are around 10 seconds, and each video is tagged with one or more video event label. 
Second, {LLP} \cite{tian2020unified} is a recent audio-visual video event dataset, which includes 11,849 video clips of 25 video categories (e.g. {\em car} and {\em cat}). Each video is 10 seconds and tagged with one or more than one video event label. 
Finally, we construct a new Audio-Visual Recognition (referred as {AVR}) dataset by unifying the AVE and LLP datasets into one dataset to obtain a large class space. AVR contains 15,992 videos of 43 video categories. We summarize the class spaces and statistics in the supplementary. 
For all the datasets, we use only the video event labels for training.

We adopt the above datasets as they contain the human annotated temporal bounding boxes that indicate when the predicted sounds occur. For instance, a temporal bounding box ``$[0,4]$'' means the sounds occur from seconds 0 to 4, which can be converted to a binary mask to evaluate whether the temporal attention values are in line with human annotations. We detail this evaluation in Section \ref{sec:analysis}.

\myparagraph{Evaluation Metrics.} 
Following the common practice for evaluating multi-label multi-class classification \cite{wu2017unified}, we adopt the following three evaluation metrics. 
Top-1 accuracy measures the fraction of instances whose most confident label is relevant. 
Mean average precision ({mAP}) is the average fraction of relevant labels ranked higher than 
other labels. 
F-score is the F-measure averaged over the instances, which is commonly adopted in sound event recognition \cite{mesaros2016metrics}. As computing the F-score requires binary predictions, we convert the model predictive scores (${\in}[0,1]$) to binary values by a threshold of 0.5. For all the metrics, higher percentages (\%) indicate better model performance. 

\myparagraph{Implementation Details.}
{We use a VGGish network \cite{hershey2017cnn} pre-trained on the YouTube dataset \cite{abu2016youtube} to extract embeddings for each audio segment (1 second).}
We use a ResNet-152 \cite{he2016deep} pre-trained on ImageNet \cite{deng2009imagenet} to extract a visual feature map that encodes each video segment. 
To ensure fair comparison, we use the Adam optimizer \cite{kingma2014adam} with the same learning rate schedule and random seed for all the methods that we compare to. 
More implementation details are given in the supplementary. Code will be publicly available. 

\subsection{Evaluating Audio-Visual Recognition}
\label{sec:recognition}

In this section, 
we conduct an ablation study on STAN, and compare to the state of the art of two types of models: 
(1) intrinsic explanation models and (2) audio-visual learners. 
All experiments on audio-visual recognition are conducted on three datasets: AVE, LLP and AVR, using three metrics: top-1, mAP (mean average precision) and F-score.    

{
\begin{table*}[!t]
	\centering
	\setlength{\tabcolsep}{5pt}
	\renewcommand{\arraystretch}{1.05}
	\begin{tabular}{p{2cm}|p{3.5cm}|ccc|ccc|ccc}
	    \multirow{2}{3pt}{Data} 
	    & \multirow{2}{3pt}{Method} 
		& \multicolumn{3}{c|}{AVE} 
		& \multicolumn{3}{c|}{LLP} 
		& \multicolumn{3}{c}{AVR} \\ 
		& & top-1 & mAP & F-score
		& top-1 & mAP & F-score
		& top-1 & mAP & F-score \\ 
		\hline 
		\multirow{2}{3pt}{Audio} 
		& audio baseline & 
         67.7 & 66.8 & 40.1 &
         79.1 & 65.5 & 60.3 &
         72.3 & 58.0 & 51.7 \\
    	& STAN (audio) & 
    	 74.4 & 77.0 & 68.2 &
     	81.8 & 72.3 & 69.1 &
     	76.5 & 67.1 & 64.1 
    	\\ \hline
    	\multirow{2}{3pt}{Vision} 
    	& visual baseline & 
        77.4 & 83.7 & 71.4 &
         61.1 & 47.4 & 46.1 &
         64.1 & 58.5 & 47.2 \\
        & STAN (visual) & 
         79.4 & 84.0 & 72.5 &
         70.8 & 59.4 & 54.0 &
         71.4 & 64.7 & 54.1 
	    \\ \hline
	    \multirow{1}{3pt}{Audio+Vision} 
	   & STAN (audio-visual) & 
	    \bf 91.8 & \bf 93.6 & \bf 85.3 &
        \bf 87.2 & \bf 82.6 & \bf 75.5 &
        \bf 86.2 & \bf 83.1 & \bf 74.5  \\
	\end{tabular}
	\vskip 0.2em
	\caption{Ablation study of STAN for audio-visual video event recognition on the AVE, LLP and AVR datasets, using the evaluation metrics: top-1, mAP (mean average precision) and F-score.
	}
	\label{tab:ablation}
\end{table*}
}
{
\begin{table*}[!t]
	\centering
	\setlength{\tabcolsep}{5pt}
	\renewcommand{\arraystretch}{1.1}
	\begin{tabular}{p{2cm}|p{3.5cm}|ccc|ccc|ccc}
	    \multirow{2}{3pt}{Model} 
	    & \multirow{2}{5pt}{Method} 
		& \multicolumn{3}{c|}{AVE} 
		& \multicolumn{3}{c|}{LLP} 
		& \multicolumn{3}{c}{AVR} \\ 
		& & top-1 & mAP & F-score
		& top-1 & mAP & F-score
		& top-1 & mAP & F-score \\ 
		\hline 
		\multirow{4}{3pt}{Explanation Models} 
        & CAM (visual) \cite{zhou2016learning} & 
        78.6 & 82.1 & 71.1 &
         61.0 & 45.8 & 46.1 &
         62.9 & 56.9 & 47.7 \\
		& ABN (visual) \cite{fukui2019attention} & 
        79.6 & 83.8 & 72.4 &
         63.2 & 47.4 & 45.1 &
         63.8 & 58.6 & 47.8 \\ 
        & AV-CAM & 
         85.6 & 88.3 & 79.6 &
    	 84.9 & 76.2 & 70.8 &
    	 83.4 & 78.4 & 71.1 
         \\
        & AV-ABN & 
    	 87.8 & 91.0 & 81.6 &
     	85.6 & 78.6 & 71.9 &
     	84.6 & 80.4 & 70.9 
	    \\
        \hline
        \multirow{4}{3pt}{Audio-Visual Models} 
		& Relationship \cite{santoro2017simple} & 
        85.3 & 88.4 & 78.6 &
         83.8 & 76.1 & 70.4 &
         83.2 & 77.8 & 68.9 \\
 		& Average Ensemble \cite{parekh2018weakly} & 
        86.8 & 90.0 & 62.8 &
         82.0 & 76.8 & 65.9 &
         79.6 & 76.4 & 59.5 \\
        & FiLM \cite{perez2018film} &  
        87.8 & 90.8 & 81.6 &
         85.0 & 79.6 & 71.3 &
         83.4 & 81.2 & 70.2 \\
		& Attention Fusion \cite{fayek2020large} & 
        89.3 & 93.2 & 81.8 &
         86.3 & 80.4 & 72.2 &
         84.8 & 81.8 & 72.6 \\
        \hline 
		Ours & STAN & 
        \bf 91.8 & \bf 93.6 & \bf 85.3 &
        \bf 87.2 & \bf 82.6 & \bf 75.5 &
        \bf 86.2 & \bf 83.1 & \bf 74.5  \\
	\end{tabular}
	\vskip 0.2em
	\caption{Evaluating state of the art intrinsic explanation models, audio-visual learners and STAN for audio-visual video event recognition 
    on the AVE, LLP and AVR datasets, using the evaluation metrics: top-1, mAP (mean average precision) and F-score.
	}
	\label{tab:recognition}
\end{table*}
}

\myparagraph{Baselines.} 
We ablate our STAN model to evaluate the effects of audio and visual data on our model.
Our {audio baseline} is a unimodal baseline that trains an MLP and a classifier upon the audio features. 
Our {visual baseline} is also a unimodal baseline that trains a convolutional layer and a classifier upon the visual features.
Furthermore, our {STAN (audio)} is an ablation of our model that uses only audio features and learns only the time attention; 
while our {STAN (visual)} is also an ablation of our model that uses only visual features and learns space-time attention. 
Finally, {STAN (audio-visual)} is our full model that integrates audio and visual data to learn space-time attention.

\myparagraph{Ablation Study.} 
Table \ref{tab:ablation} shows our ablative evaluation on audio-visual video event recognition. 
As can be seen, compared to the audio baseline and visual baseline that learn from one single modality, STAN (audio) and STAN (visual) greatly boost the model performance by learning the explainable attention on the audio or visual data. For instance, when comparing the audio baseline and STAN (audio), the F-score is increased from 40.1/60.3/51.7 to 68.2/69.1/64.1 on the AVE/LLP/AVR datasets. Similarly, when comparing the visual baseline and STAN (visual), the F-score is improved from 71.4/46.1/47.2 to 72.4/54.0/54.1 on AVE/LLP/AVR. 
Compared to the unimodal models, our STAN (audio-visual) further enhances the model performance. The F-score of STAN (audio-visual) is raised upon STAN (visual) from 72.4/54.0/54.1 to 85.3/75.5/74.5 on AVE/LLP/AVR. 
The similar improving trends can also be observed in other evaluation metrics including top-1 and mAP. 
Overall, these results suggest the joint benefits of composing the audio and visual modalities and learning the explainable space-time attention. In the following, we refer to our full model STAN (audio-visual) as STAN. 

\myparagraph{Compared Methods.} 
We compare STAN to two groups of models: (1) intrinsic explainable models that incorporate explanation as an intrinsic model component; (2) multi-modal representation learners that integrate audio and visual data. 
We describe these two groups of models as below.

\begin{itemize}[noitemsep,topsep=2pt,itemsep=1pt,leftmargin=8pt]
\item {CAM} \cite{zhou2016learning} (class activation maps): A visual explanation model that learns to localize the discriminative regions.

\item {ABN} \cite{fukui2019attention} (attention branch network): An attention-based visual explanation model that learns visual attention maps to pinpoint class-agnoistic discriminative regions. 

\item {AV-CAM}: We extend the vanilla CAM to learn upon the concatenation of visual feature maps and audio features. 

\item {AV-ABN}: Same as AV-CAM, we extend the vanilla ABN to learn from audio and visual data. 
\end{itemize}

\remark{Remark.} 
CAM is used as a model component in the above methods. 
However, our model especially learns class activation values (CAV) along time, which allows to uncover both the spatial and temporal dynamics in videos.

\begin{itemize}[noitemsep,topsep=2pt,itemsep=1pt,leftmargin=8pt]
\item {Relationship} \cite{santoro2017simple}: 
A multi-modal module. 
It takes in the visual features and 
audio features, 
followed by concatenation and an MLP 
to fuse the audio-visual features. 

\item {Average Ensemble} \cite{parekh2018weakly}:
An ensemble of the unimodal audio and visual baselines 
by averaging their predictions. 

\item {FiLM} \cite{perez2018film}: 
A Feature-wise Linear Modulation module, 
where the audio features are used 
to modulate the visual features
through a learnable affine transformation. 

\item {Attention Fusion} \cite{fayek2020large}: 
A recent state-of-the-art audio-visual feature learner  
that fuses the audio and visual features by 
an attention-based weighted average. 

\end{itemize}

\remark{Remark.} 
Relationship and FiLM are two representative multi-modal models proposed to learn from images and texts. We adapt these two methods by replacing the text features with audio features, and use temporal average pooling to obtain one audio-visual feature per video. As Average Ensemble and Attention Fusion are both designed for audio-visual learning, we apply them to our context directly. 

\myparagraph{Comparative Results.} 
Table \ref{tab:recognition} shows the comparative evaluation on two groups of models and our STAN on audio-visual video event recognition. 
Among the group of intrinsic explanation models including CAM, ABN, AV-CAM and AV-ABN, we can observe the benefits of learning from audio-visual data jointly, e.g., the performance of CAM and ABN are improved substantially in their multi-modal variants. However, compared to the best model AV-ABN in this group, our STAN still outperforms AV-ABN significantly, increasing the F-score from 81.6/71.9/70.9 to 85.3/75.5/74.5 on AVE/LLP/AVR respectively. The performance advantages of STAN over AV-CAM and AV-ABN indicate the synergistic merits of learning the space attention based on class activation maps and the time attention based on class activation values. By design, the space-time attention also selectively activates the space-time audio-visual features (Eq. \eqref{eq:feature}) to facilitate better audio-visual learning. 

In the group of audio-visual learners, we see that STAN still outperforms the best model Attention Fusion, improving the F-score from 81.8/72.2/72.6 to 85.3/75.5/74.5 on AVE/LLP/AVR respectively. Although Attention Fusion also learns attention to selectively fuse the two modalities, STAN learns a more advanced space-time attention that can select the essential audio and visual information both over space and along time, thus offering superior model performance for audio-visual video event recognition. 

\subsection{Evaluating Audio-Visual  Explanations}
\label{sec:analysis}

To study how our explainable attention relates to model explanations and human explanations, we design two experiments to analyze its explainablity: 
(1) robustness analysis via perturbation tests, and 
(2) pointing games on localizing sounds. 
The former aims to uncover whether the attention value provides the importance of each feature towards model prediction; 
while the latter focuses on examining whether the attention is in line with human annotations. 

\begin{figure}[!t]
\includegraphics[width=0.478\textwidth]{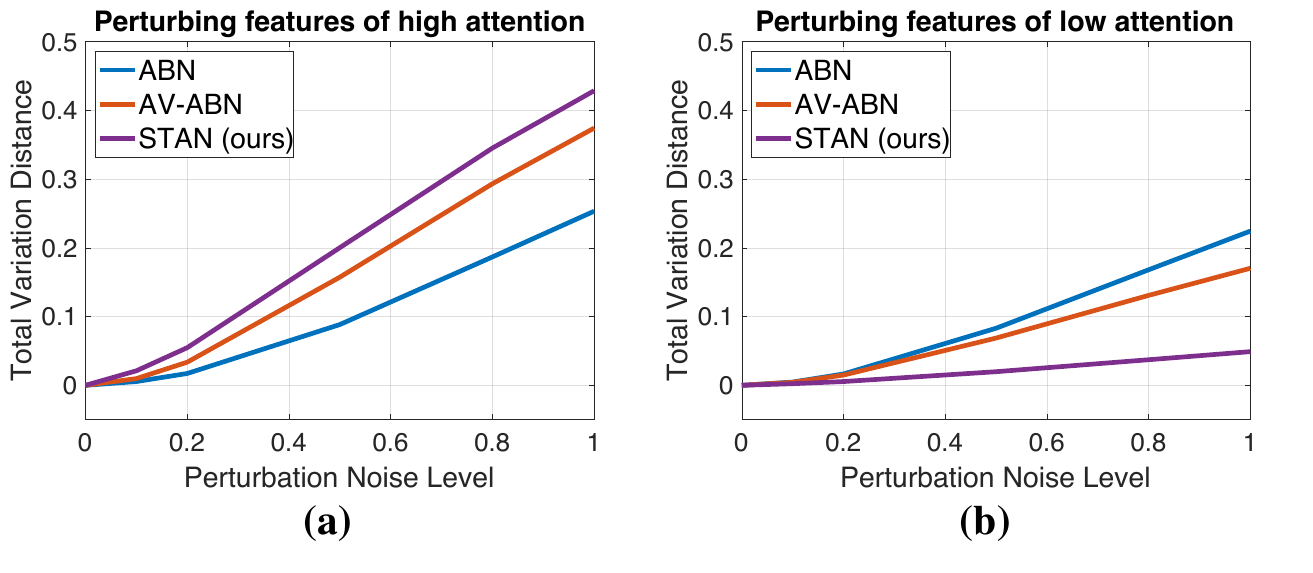}
\caption{
Perturbation tests on {\bf (a)} relevant features of high attention values; 
{\bf (b)} irrelevant features of low attention values. 
Note: Increasing the perturbation noise level on features of high/low attention values should lead to higher/lower TVD.
Dataset: AVR. 
}
\label{fig:perturbation}
\end{figure}

\myparagraph{Perturbation Tests.} 
Here, we examine if our learned attention-based explanations uncover the importance of each feature towards the model's decision. We conduct perturbation tests following the evaluation criteria in robustness analysis for evaluating model explanations \cite{hsieh2020evaluations}. 
Specifically, our perturbation tests are driven by two evaluation criteria. First, perturbing the relevant features (i.e., features with high attention scores or importance values) should lead to an appreciable change in the model predictions. Second, perturbing the irrelevant features (i.e., features with low attention scores or importance values) should {\em not} lead to an appreciable change in model predictions. Based on these criteria, we add perturbation noise to the features, following a Gaussian distribution $\eta \sim \mathcal{N}(0, \sigma^{2})$, where $\sigma^{2}$ is increased from 0 to 1 to vary the level of noise. 
As the attention scores lie in $[0,1]$, we count the features with higher attention scores (${\geqslant}0.5$) as relevant, and the features with lower attention scores (${<}0.5$) as irrelevant. To quantify the changes in the model predictions, we measure the total variation distance (TVD) between two probability distributions P and Q derived before and after adding perturbation noise, where 
$\text{TVD} (P, Q) = \frac{1}{2} \sum_{i=1}^{K}|P_i-Q_i|$. 

Figure \ref{fig:perturbation} presents the results of perturbation tests on the AVR dataset for two types of features: (a) relevant features that have higher attention values; and (b) irrelevant features with lower attention values. We compare our STAN to two strong intrinsic explainable models: ABN and AV-ABN (described in Section \ref{sec:recognition}), which also learn attention maps to locate the essential features. For each level of perturbation noise, we randomly sample the 100 different noises and compute the mean TVD of 100 tests. We give an algorithm overview of our perturbation tests in the supplementary. 

On the one hand, as Figure \ref{fig:perturbation} (a) shows, when increasing the perturbation noise level in relevant features, the TVD is increased among all the methods. However, the changes are much larger in STAN as compared to those of ABN and AV-ABN. This means that the features of high attention values predicted by STAN correspond to the regions essential towards model predictions. On the other hand, as Figure \ref{fig:perturbation} (b) shows, when increasing the perturbation noise level in irrelevant features, the predictive changes in STAN are almost imperceptible compared to the much bigger changes in ABN and AV-ABN. This indicates the features of low attention values in STAN are indeed not important for making model predictions. In summary, these results indicate that our model offers explainable attention that picks up the features of importance for model decision, and serves as a good proxy of model explanations.

\begin{figure*}[!t]
\includegraphics[width=0.98\textwidth]{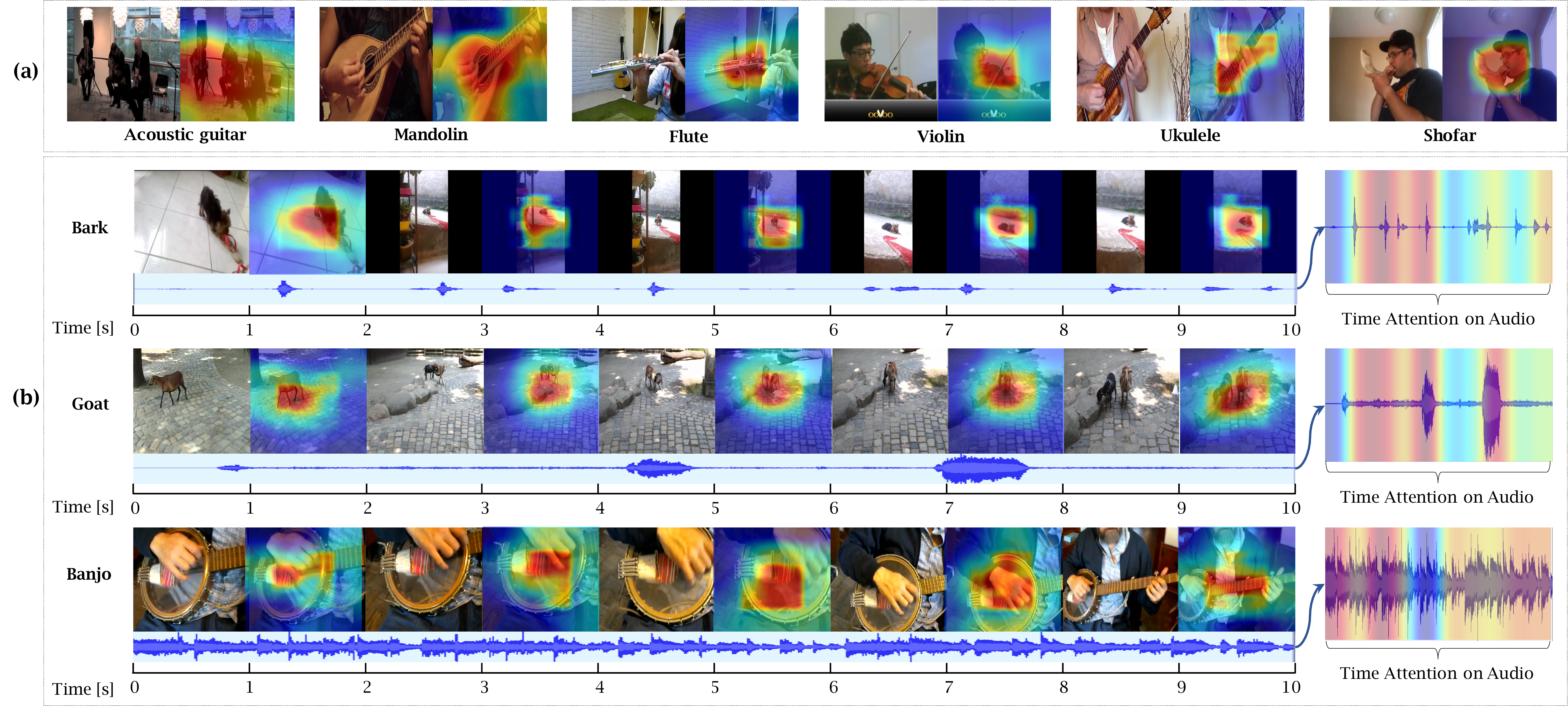}
\caption{
Visualizing attention maps with STAN. 
(a) Space attention on image frames from different video classes. 
(b) Space attention maps over different image frames in the videos and the time attention on the co-occurring audios. Dataset: AVR. 
}
\label{fig:attention}
\end{figure*}

{
\begin{table}[!t]
	\centering
	\setlength{\tabcolsep}{3pt}
	\renewcommand{\arraystretch}{1.05}
	\begin{tabular}{l|c|c|c}
	    Method & AVE & LLP & AVR \\
		\hline 
		STAN (audio)+soft attention 
		& 0.47 & 0.42 & 0.44
		\\
		STAN (audio-visual)+soft attention
		& \bf 0.46 & \bf 0.40 & \bf 0.37
		\\ \hline
		STAN (audio)+binary attention
		& 0.33 & \bf 0.23 & 0.25
		\\
		STAN (audio-visual)+binary attention
		& \bf 0.29 & 0.24 & \bf 0.20
		\\
	\end{tabular}
	\vskip 0.2em
	\caption{Evaluation of pointing games for localizing sounds on AVE, LLP and AVR. Metric: MAE (the lower the better). 
	}
	\label{tab:point}
\end{table}
}
\myparagraph{Pointing Games on Localizing Sounds.} 
To further evaluate how our attention-based explanations relate to human explanations, we conduct pointing games using available human annotations on audio.
This evaluation shares a similar spirit as the pointing games for localizing visual objects in images \cite{zhang2018top}. However, we design this test in the audio domain to evaluate the temporal localization ability of sounds. We exploit the human-annotated temporal bounding boxes that indicate when the sounds occur, as described in Section \ref{sec:setup}. 
Hence, they serve as a proxy of human explanations. 

We compare the time attention to the binary masks obtained by converting a human annotation to a grounding mask: $[0,...,1,1,0]$, where $1$ refers to the time when the sounds occur. We compute the mean absolute error (MAE) between the time attention and the binary mask, which measures the difference between our explanation and the human explanation. 
We evaluate STAN in two modes, i.e. STAN (audio) and STAN (audio-visual). As described in Section \ref{sec:recognition}, these are the only two explanation models that offer the time attention on audio. We also evaluate soft attention obtained from the model and binary attention derived by binarizing the soft attention with a threshold of 0.5. 

As Table \ref{tab:point} shows, STAN (audio-visual) overall offers the lower MAE compared to STAN (audio) on two types of attentions across three different datasets. Notably, when using soft attention, STAN (audio-visual) obtains a lower MAE of 0.46/0.40/0.37 compared to 0.47/0.42/0.44 by STAN (audio) on AVE/LLP/AVR. This indicates that STAN (audio-visual) has a better temporal localization ability of sounds. In other words, the time attention learned by STAN (audio-visual) is closer to the groundtruth human annotations.

\myparagraph{Visualizing Attention Maps.}
We visualize the attention maps of STAN in Figure \ref{fig:attention}. As Figure \ref{fig:attention} (a) shows, the space attention maps of different video classes cover the most discriminative visual regions. For instance, in the video {\em Flute} and {\em Violin}, the flute and violin are highlighted by STAN. This means that the space attention can pick up the class-relevant visual cues. 
In Figure \ref{fig:attention} (b), we can observe that the space attention on video and the time attention on audio work in different but synergistic ways. 
For instance, in the video {\em Goat}, the space attention operates on the image frames to locate the moving goats in the video; while the time attention operates on the audio to pick up the relevant audio cues along time. 
Overall, these results confirm that our formulation of space and time attention works synergistically to discover the informative audio and visual cues over space and time, which also resembles how humans interpret video events with both audio and visual content. 

\section{Conclusion}

We presented a novel intrinsic explanation model for audio-visual recognition, which sheds light on a new aspect for explainable modeling on multi-modal data collected over space and time. 
Our proposed space-attention network (STAN) first composes the audio and visual features, and learns attention upon class activation maps over space and class activation values along time. 
The learned space and time attention maps can be considered as explanations on the visual and audio modalities for video event recognition.
Our comprehensive experiments demonstrate that STAN is a strong audio-visual representation learners and offers impressive model performance on audio-visual event recognition. Our analysis on explainablity also shows that STAN provides meaningful explanations that are closely in line with model explanations and human explanations. 

\myparagraph{Acknowledgements} This work has been partially funded by the ERC (853489 - DEXIM) and by the DFG (2064/1 – Project number 390727645).

{\small
\bibliographystyle{ieee_fullname}
\bibliography{reference}
}

\newpage
\section*{\Large Supplementary}
\setcounter{figure}{0}
\setcounter{section}{0}
\setcounter{table}{0}
\setcounter{algorithm}{0}
\renewcommand\thesection{\Alph{section}}
\renewcommand\thefigure{\Alph{figure}}
\renewcommand\thetable{\Alph{table}}
\renewcommand\thealgorithm{\Alph{algorithm}}

We provide additional details about the datasets in Section \ref{sec:supp_datasets}, more implementation details of our proposed model STAN in Section \ref{sec:supp_implementation}, and the algorithm descriptions for the perturbation tests in Section \ref{sec:supp_perturbation}. Further, we present additional qualitative results in Section \ref{sec:supp_results}.

\section{Datasets}\label{sec:supp_datasets}

Tables \ref{tab:AVE}, \ref{tab:LLP} and \ref{tab:AVR} summarize the class label space of the three audio-visual datasets: AVE, LLP and AVR, which include 28, 25 and 43 different classes respectively. In Table \ref{tab:statistic}, we summarize the statistics of the training, validation and test set.

{
\setlength{\tabcolsep}{2pt}
\renewcommand{\arraystretch}{1.1}
\begin{table}[!ht]
   \footnotesize
	\begin{tabular}{l|l}
	\hlineB{2}
	\bf Hyper Class & \bf Class Names \\
	\hline
	Animal & Bark; Cat; Goat; Horse; Rodents, rats, mice \\
    \hline
	\multirow{2}{3pt}{Instrument} & Accordion; Acoustic guitar; Banjo; Flute; \\
	& Mandolin; Shofar; Ukulele; Violin, fiddle \\
	\hline
	Home & Clock; Frying (food); Toilet flush \\
	\hline
	\multirow{2}{3pt}{Human} & Baby cry, infant cry; Female speech, woman speaking; \\
	& Male speech, man speaking \\
	\hline
	\multirow{2}{3pt}{Vehicle} & Bus; Fixed-wing aircraft, airplane; Helicopter;  \\
	& Motorcycle; Race car, auto racing; Train horn; Truck \\
	\hline
	Others & Chainsaw; Church bell \\
	\hlineB{2} 
	\end{tabular}
	\vskip 0.1em
	\caption{The audio-visual class space of the AVE dataset.} 
	\label{tab:AVE}
\end{table}
}

{
\setlength{\tabcolsep}{2pt}
\renewcommand{\arraystretch}{1.1}
\begin{table}[!ht]
   \footnotesize
	\begin{tabular}{l|l}
	\hlineB{2}
	\bf Hyper Class & \bf Class Names \\
	\hline
	Animal & Cat; Chicken rooster; Dog \\
	\hline
	Instrument & Accordion; Acoustic guitar; Banjo; Cello; Violin, fiddle \\
	\hline
	\multirow{2}{3pt}{Home} & Blender; Frying (food); Lawn mower; \\
	& Telephone bell ringing; Vacuum cleaner \\
	\hline
	\multirow{2}{3pt}{Human} & Baby cry, infant cry; Baby laughter; \\
	& Cheering; Clapping; Singing; Speech \\
	\hline
	Vehicle & Car; Helicopter; Motorcycle \\
	\hline
	Others & Basketball bounce; Chainsaw; Fire alarm \\
	\hlineB{2} 
	\end{tabular}
	\vskip 0.1em
	\caption{The audio-visual class space of the LLP dataset.} 
	\label{tab:LLP}
\end{table}
}

{
\setlength{\tabcolsep}{2pt}
\renewcommand{\arraystretch}{1.1}
\begin{table}[!ht]
   \footnotesize
	\begin{tabular}{l|l}
	\hlineB{2}
	\bf Hyper Class & \bf Class Names \\
	\hline
	\multirow{2}{3pt}{Animal} & Bark (Dog); Cat; Chicken rooster; Goat; Horse; \\
	& Rodents, rats, mice \\
	\hline
	\multirow{2}{3pt}{Instrument} & Accordion; Acoustic guitar; Banjo; Cello; \\
	& Flute; Mandolin; Shofar; Ukulele; Violin, fiddle \\
	\hline
	\multirow{2}{3pt}{Home} & Blender; Clock; Frying (food); Lawn mower; \\
	& Telephone bell ringing; Toilet flush; Vacuum cleaner \\
	\hline
	\multirow{3}{3pt}{Human} & Baby cry, infant cry; Baby laughter; Cheering; Clapping; \\
	& Female speech, woman speaking; \\ 
	& Male speech, man speaking; Singing; Speech \\
	\hline
	\multirow{2}{3pt}{Vehicle} & Car; Bus; Fixed-wing aircraft, airplane; Helicopter; \\
	& Motorcycle; Race car, auto racing; Train horn; Truck \\
	\hline
	Others & Basketball bounce; Chainsaw; Church bell; Fire alarm \\
	\hlineB{2} 
	\end{tabular}
	\vskip 0.1em
	\caption{The audio-visual class space of the AVR dataset.} 
	\label{tab:AVR}
\end{table}
}

{
\setlength{\tabcolsep}{2pt}
\renewcommand{\arraystretch}{1}
\begin{table}[!ht]
   \small
    \centering
	\begin{tabular}{c|c|c|c|c}
	\hlineB{2}
	\bf Dataset & \bf \# Total & \bf \# Training & \bf \# Val & \bf \# Test \\ \hline
    AVE & 4,143 & 3,339 & 402 & 402 \\
    LLP & 11,850 & 10,000 & 650 & 1,200 \\
    AVR & 15,993 & 13,339 & 1,602 & 1,052 \\
	\hlineB{2} 
	\end{tabular}
	\vskip 0.1em
	\caption{Dataset statistics for the AVE, LLP, and AVR datasets.} 
	\label{tab:statistic}
\end{table}
}

\section{Implementation Details}\label{sec:supp_implementation}

As aforementioned in Section 4.1 in the main paper, the audio and visual features are extracted from a VGGish network \cite{hershey2017cnn} pre-trained on YouTube-8M \cite{abu2016youtube} and a ResNet-152 pre-trained on ImageNet \cite{deng2009imagenet}. For each 10 second video (i.e., $T=10$), audio features (with size $T \times D_a$) and visual feature maps (with size $T \times H \times W \times D_c$) are extracted to encode the audio and video segment per second. For pre-trained networks, we empirically find that a pre-trained image encoder works better than a video encoder pre-trained on a video dataset such as Kinetics \cite{carreira2017quo}. This is likely caused by the fact that most audio-visual video events do not contain a lot of motion information for actions, while an ImageNet pre-trained network offers more generic visual features than the visual features extracted from a pre-trained video network. The audio and visual features, extracted from pre-trained networks, further serve as the inputs to STAN. 
To understand the learned attention maps, we show the space attention maps on image frames and the time attention maps on raw audio data. 
Since the convolutions in the image encoder reduce the spatial sizes of the visual features, we resize the space attention maps to match the original image size. Similarly, the raw audio time attention maps are of small resolution. Hence, we upsample the attention maps to a larger size for better visual quality. 
PyTorch code will be released publicly. 

\section{Algorithm for Perturbation Tests}\label{sec:supp_perturbation}

We give an algorithm overview for a general perturbation test (as mentioned in Section \ref{sec:analysis} in the main paper) in Algorithm~\ref{alg:perturb-general}, which either adds input noise ($\eta \sim \mathcal{N}(0, \sigma^{2})$) to the relevant features (i.e. features with higher attention values: $\mathbbm{1}_{[{A} \geq 0.5]}$) or the irrelevant features (i.e. features with lower attention values: $\mathbbm{1}_{[{A} < 0.5]}$). The robustness under perturbation is measured as the total variation distance (TVD) between the predictive distributions before and after perturbation:   
$\text{TVD} (\hat{y}, \hat{y}^p) = \frac{1}{2} \sum_{i=1}^{K}|\hat{y}-\hat{y}^p|$. 
We also detail the perturbation test for STAN in Algorithm~\ref{alg:perturb-stan}, which adds perturbation noise to the audio and visual features similar to Algorithm~\ref{alg:perturb-general}. Note that for each input, {we average the TVD values over 100 random perturbations}.

\begin{algorithm}[!t]
\begin{algorithmic}[1]

\State $\bm{x} \gets \mathcal{F}_{\textrm{feat}}(\textbf{X})$, ${A} \gets \mathcal{F}_{\textrm{attn}}(\bm{x})$, $\hat{y} \gets \mathcal{F}_{\textrm{classify}}(\bm{x}, {A})$ 

\If{\text{perturb relevant features}} 
    \State $\epsilon = \mathbbm{1}_{[{A} \geq 0.5]}$
    \Comment{{\small mask on input of high attention}}
\Else
    \State $\epsilon = \mathbbm{1}_{[{A} < 0.5]}$
    \Comment{{\small mask on input of low attention}}
\EndIf

\For{$p \gets 1 \textrm{ to } 100$}
    \State $\eta \sim \mathcal{N}(0, \sigma^{2})$ 
    \Comment{{\small sample random input noise}}
    
    \State $\bm{x}^p \gets \bm{x} + \eta * \epsilon$ 
    \Comment{{\small add noise to input features}}
    
    \State $\hat{y}^p \gets \mathcal{F}_{\textrm{classify}}(\bm{x}^p, {A})$ 
    \Comment{{\small forward perturbed input}}
    
    \State $\Delta\hat{y}^p \gets \text{TVD}(\hat{y}^p, \hat{y})$ 
    \Comment{{\small compute TVD}}
\EndFor

\State $\Delta\hat{y}^{mean} \gets \textrm{Mean}_p(\Delta\hat{y}^p)$
\Comment{{\small average TVD}}

\end{algorithmic}

\caption{General perturbation test on input features.}
\label{alg:perturb-general}
\end{algorithm}

\begin{algorithm}[!t]
\begin{algorithmic}[1]

\State $\bm{a} \gets \mathcal{F}_{\textrm{featAudio}}(\textbf{X}_{a})$, $\bm{v} \gets \mathcal{F}_{\textrm{featVideo}}(\textbf{X}_{v})$

\State ${A}^{\bm{s}} \gets \mathcal{F}_{\text{space-branch}}(\bm{a}, \bm{v})$,
${A}^{\bm{t}} \gets \mathcal{F}_{\text{time-branch}}(\bm{a}, \bm{v})$

\State ${A}^{\bm{st}} \gets {A}^{\bm{s}} \otimes {A}^{\bm{t}}$, $\bm{x}^{\bm{st}} \gets \mathcal{F}_{\text{space-time}}(\bm{a}, \bm{v})$

\State $\hat{y} \gets \mathcal{F}_{\textrm{classify}}(\bm{x}^{\bm{st}}, {A}^{\bm{st}})$ 

\If{\text{perturb relevant features}} 
    \State $\epsilon = \mathbbm{1}_{[{A} \geq 0.5]}$
    \Comment{{\small mask on input of high attention}}
\Else
    \State $\epsilon = \mathbbm{1}_{[{A} < 0.5]}$
    \Comment{{\small mask on input of low attention}}
\EndIf

\For{$p \gets 1 \textrm{ to } 100$}
    \State $\eta \sim \mathcal{N}(0, \sigma^{2})$ 
    \Comment{{\small sample random input noise}}
    
    \State $\bm{a}^p \gets \bm{a} + \eta * \epsilon$, 
    $\bm{v}^p \gets \bm{v} + \eta * \epsilon$ 
    \Comment{{\small add noise}}
    
    \State $\bm{x}^{\bm{st},p} \gets \mathcal{F}_{\text{space-time}}(\bm{a}^p, \bm{v}^p)$
    \Comment{{\small forward propagation}}
    
    \State $\hat{y}^p \gets \mathcal{F}_{\textrm{classify}}(\bm{x}^{\bm{st},p}, {A}^{\bm{st}})$ 
    \Comment{{\small forward perturbed input}}
    
    \State $\Delta\hat{y}^p \gets \text{TVD}(\hat{y}^p, \hat{y})$
    \Comment{{\small compute TVD}}
    
\EndFor

\State $\Delta\hat{y}^{mean} \gets \textrm{Mean}_p(\Delta\hat{y}^p)$
\Comment{{\small average TVD}}

\end{algorithmic}
\caption{STAN perturbation test on input features.}
\label{alg:perturb-stan}
\end{algorithm}

\section{Additional Results}\label{sec:supp_results}

\begin{figure}[!t]
\includegraphics[width=0.478\textwidth]{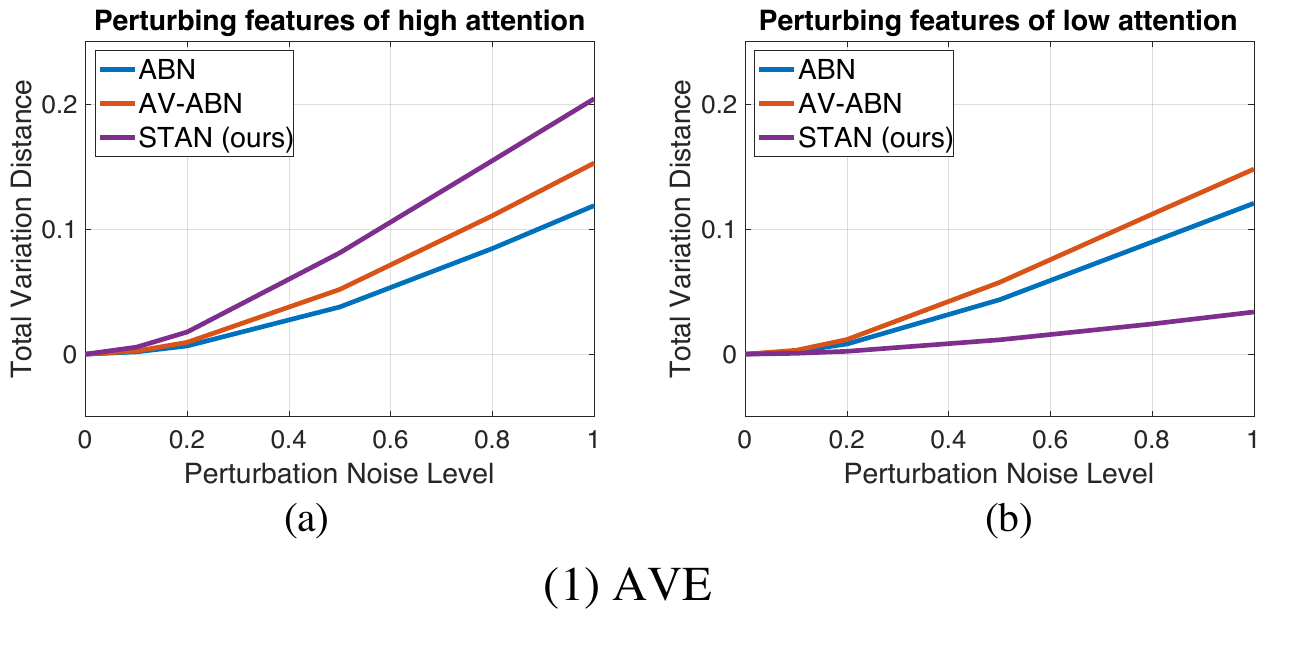}
\includegraphics[width=0.478\textwidth]{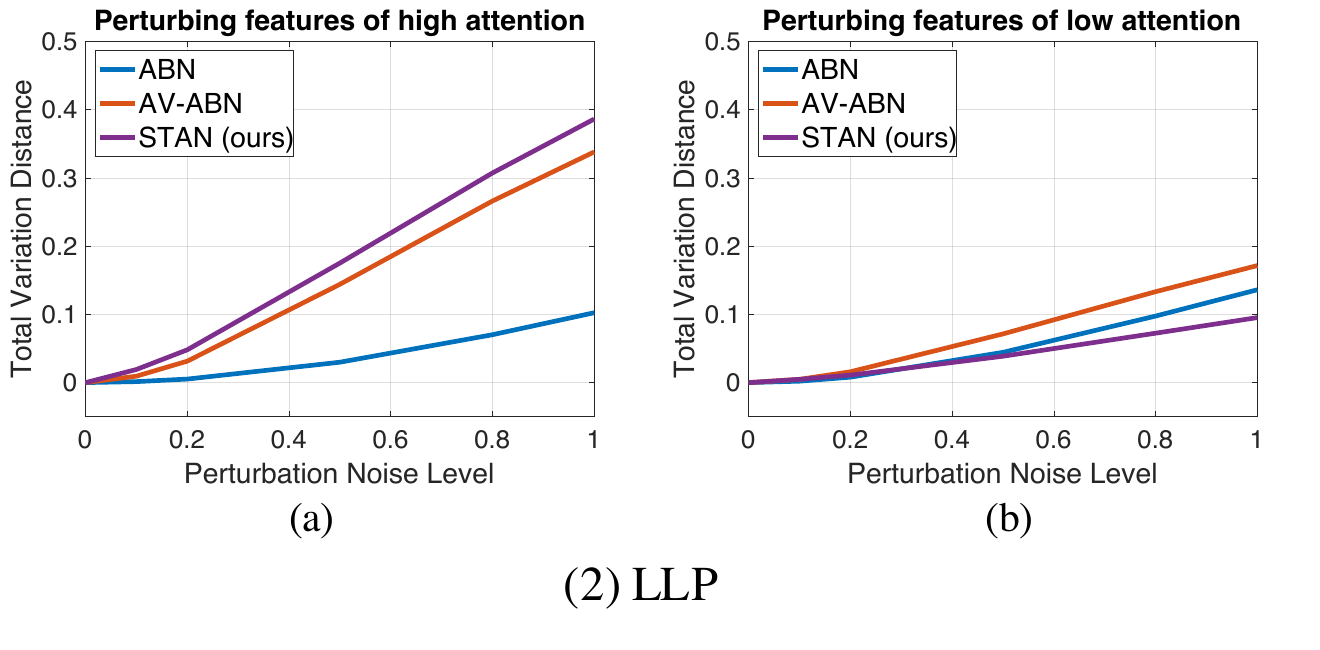}
\caption{
Perturbation tests on the AVE and LLP datasets for {(a)}~relevant features with high attention values, and 
{(b)}~irrelevant features with low attention values. 
Note: Increasing the perturbation noise level on features with high / low attention values should lead to higher / lower TVD.
}
\label{fig:perturbation-sup}
\end{figure}

\begin{figure*}[!t]
\includegraphics[width=0.98\textwidth]{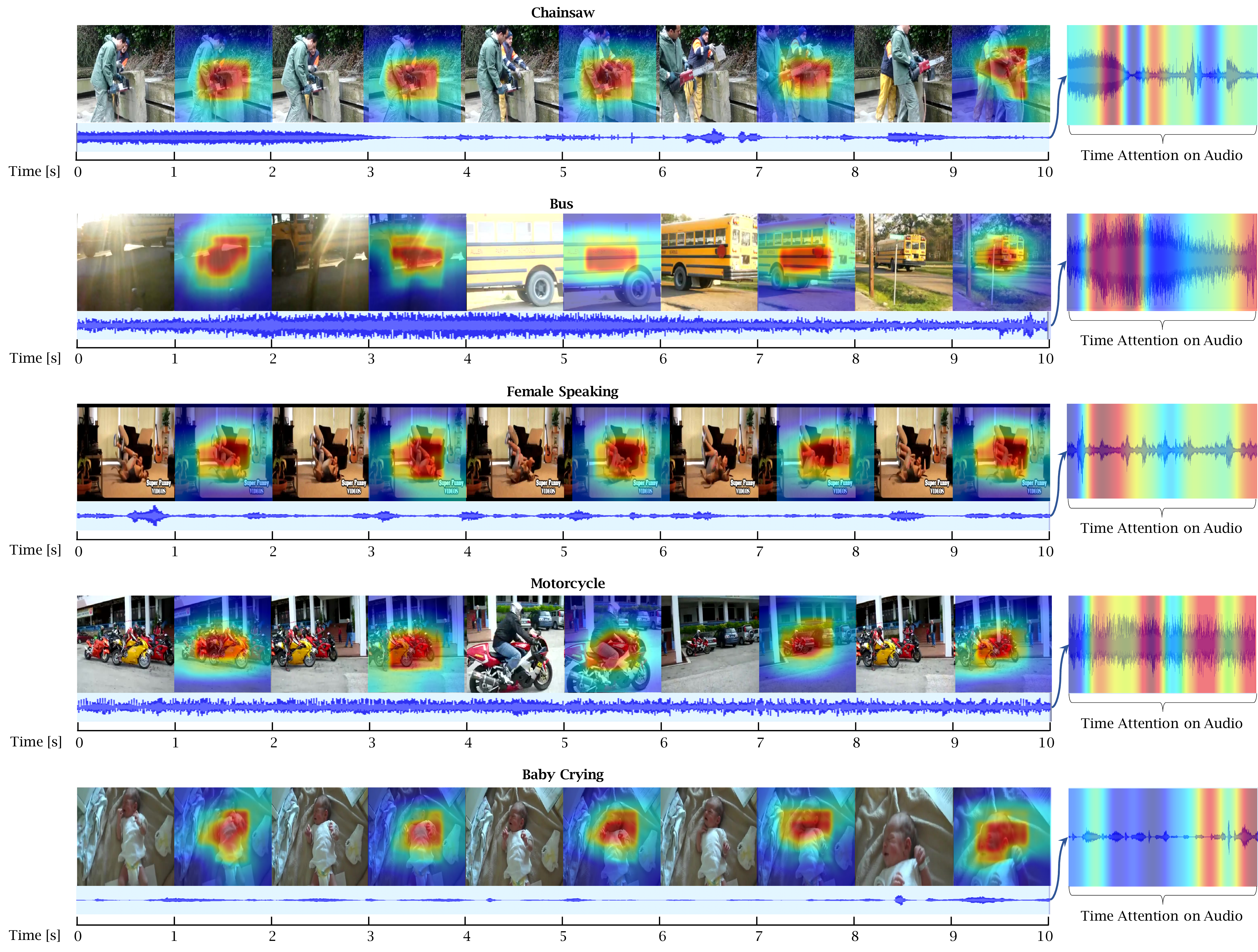}
\caption{
Visualizing attention maps with STAN, which shows
space attention maps over different image frames in the videos and the time attention maps on the co-occurring audio waveform files. Dataset: AVR.
}
\label{fig:attention-sup}
\end{figure*}

\myparagraph{Additional Results on Perturbation Tests.} 
Figure \ref{fig:perturbation-sup} shows the perturbation tests on the AVE and LLP datasets. Similar to Figure \ref{fig:perturbation} in the main paper that shows perturbation tests on the AVR dataset, we can observe the same trends in Figure \ref{fig:perturbation-sup}. 
As Figure \ref{fig:perturbation-sup} (1)-(a) and (2)-(a) show, when increasing the perturbation noise level in relevant features, the TVD is increased among all the methods. The changes in predictions (as quantified by TVD) are much larger for STAN than the changes made by ABN and AV-ABN. As Figure \ref{fig:perturbation-sup} (1)-(b) and (2)-(b) show, when increasing the perturbation noise level in irrelevant features, the predictive changes in STAN are much smaller compared to the much bigger changes made by ABN and AV-ABN. 
These observations again confirm that the features with high attention values predicted by STAN correspond to the regions essential for model predictions; while the features with low attention values in STAN are less essential for making model predictions. In other words, our model offers explainable attention which picks up on the features of importance for making the model decision.

\myparagraph{Additional Results on Attention Visualization.} 
Similar to Figure 4 in the main paper, we show more visual examples of the space and time attention maps obtained from our STAN in Figure \ref{fig:attention-sup}. As can be seen, the space and time attention maps work in different ways to pick up the discriminative visual and audio cues over space and time. As shown in the main paper, our STAN not only offers meaningful attention-based explanations, but also provides better model performance in audio-visual video event recognition.

\end{document}